\documentclass[runningheads]{llncs}

\usepackage[mobile]{eccv}

\usepackage{eccvabbrv}
\usepackage{graphicx}
\usepackage{booktabs}
\usepackage{wrapfig}
\usepackage{multirow}
\usepackage{tabularx}
\usepackage{algorithm}
\usepackage{algpseudocode}
\usepackage{listings}
\usepackage{xcolor}
\usepackage[accsupp]{axessibility}
\usepackage{hyperref}
\usepackage{orcidlink}

\begin{document}

\title{Learning from Failure: Inference-Time Self-Improvement for Computer-Use Agents} 

\titlerunning{Inference-Time Self-Improvement for Computer-Use Agents}

\author{
Xueqiao Sun\inst{1,2} \and
Xiaohan Wang\inst{1} \and
Ludwig Schmidt\inst{1} \and
Serena Yeung-Levy\inst{1,\dagger} \and
Yuhui Zhang\inst{1,\dagger} 
}

\authorrunning{Sun et al.}

\institute{$^1$Stanford University, Stanford, CA 94305, USA \\ $^2$Tsinghua University, Beijing, 100084, China}

{
  \renewcommand{\thefootnote}{\fnsymbol{footnote}}
  \footnotetext[0]{$^\dagger$Corresponding authors: \texttt{\{yuhuiz,syyeung\}@stanford.edu} (equal advising).}
}

\maketitle

\begin{abstract}

Computer-use agents, which leverage multimodal large language models (MLLMs) to operate computers and complete tasks, have attracted significant attention for their utility and versatility. A major challenge in developing these agents is collecting large-scale, high-quality trajectories. The standard approach generates synthetic data through a self-improving loop: an agent is placed in a verifiable environment and iteratively fine-tuned on its successful trajectories. Despite its effectiveness, this paradigm exploits only successful trajectories and discards the failed ones, even though failures carry rich information about a model's weaknesses. In this work, we explore a complementary failure-driven self-improvement loop, a data-centric paradigm that turns failed trajectories into agent improvements. Specifically, we employ an LLM to diagnose failure modes, propose inference-time solutions, and generate code patches—lightly verified by humans—that upgrade the agent. We validate this approach with the state-of-the-art OpenCUA-72B model on the OSWorld benchmark, improving the success rate from 42.3\% to 48.9\%, a gain of 6.6 percentage points, without any additional training cost and with only modest inference overhead. Our results demonstrate that failure-driven self-improvement is a viable complement to success-based pipelines, enabling more efficient agent improvement. Code is available at \url{https://github.com/snow10072740/Learning_from_Failure}.

\end{abstract}
    
\section{Introduction}
\label{sec:intro}

\begin{figure}[t] 
  \centering
  \includegraphics[width=\linewidth]{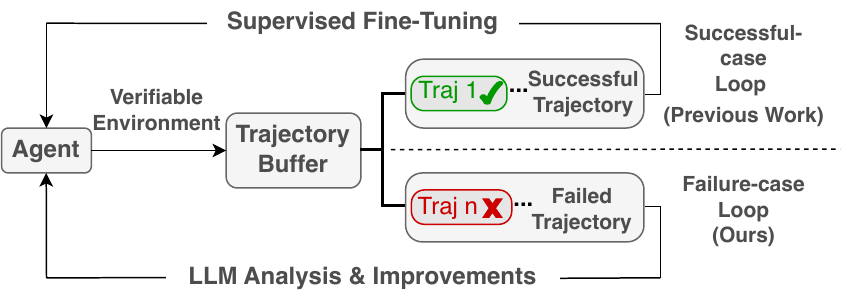} 
  \caption{Illustration of self-improving loops. While prior work focuses on fine-tuning the agent with collected successful trajectories, we explore the failure-case loop, which makes use of the large number of failure cases through LLM analysis and self-improvement.}
  \label{fig:2loop_intro}
\end{figure}

Computer-use agents—systems that employ multimodal large language models (MLLMs) to operate a computer and complete complex tasks such as buying tickets or creating slides—have recently attracted substantial attention for their practical value~\cite{wang2025opencuaopenfoundationscomputeruse,qin2025uitarspioneeringautomatedgui,agashe2024agent,agashe2025agent,yuan2025surfer2,song2025coact1,feizi2025grounding,yao2025agentic,hu2025osagents}. In such settings, an MLLM observes the current computer state (e.g., a screenshot) and predicts the next action conditioned on the task instruction, iteratively producing a trajectory of state–action pairs until the task is completed~\cite{xie2024osworld}.

A key requirement for building effective computer-use agents is the availability of high-quality trajectories~\cite{wang2025opencuaopenfoundationscomputeruse}. Although human demonstrations are the most reliable source, they are extremely costly and do not scale. Consequently, recent work relies heavily on synthetic trajectories generated in verifiable environments: an agent is placed in an environment, executes a trajectory, and an automatic verifier determines whether the trajectory is correct. Successful trajectories are added to a dataset that is used to further fine-tune the agent, forming the iterative cycle agent $\rightarrow$ environment $\rightarrow$ successful trajectory $\rightarrow$ SFT $\rightarrow$ improved agent~\cite{wang2025opencuaopenfoundationscomputeruse,qin2025uitarspioneeringautomatedgui,ye2025mobileagentv3fundamentalagentsgui}. We refer to this as the \emph{successful-case loop}, which substantially reduces reliance on expensive human annotations.

However, a major limitation of this loop is that all failed trajectories are discarded, despite the substantial cost of building verifiable environments. Although constructing verifiable environments is cheaper than collecting human trajectories, it still requires considerable human engineering effort. This raises a natural question: can we extract value from failed trajectories instead of wasting them?

In this work, we investigate how to leverage failed trajectories to help agents self-improve. We introduce a complementary loop that enhances the agent at inference time, as shown in Figure~\ref{fig:2loop_intro}. Given the environment and a collection of failed trajectories, we ask an LLM to analyze these failures, categorize common failure modes, and propose actionable solutions. We then convert these solutions into patch-like code modifications that can be injected into the trained agent's inference-time behavior, yielding an improved agent. This produces a second loop, agent $\rightarrow$ environment $\rightarrow$ failed trajectories $\rightarrow$ LLM analysis \& improvement $\rightarrow$ new agent, which we call the \emph{failure-case loop}. By delivering structured improvements as code-level updates, this loop gives agents a practical mechanism to continuously refine their behavior as they interact with the environment.

We explore this failure-case loop in OSWorld, the most widely used verifiable environment for computer-use agents. Starting from the strongest open-source model, OpenCUA-72B~\cite{wang2025opencuaopenfoundationscomputeruse}, which initially achieves 42.3\% accuracy on OSWorld, our LLM-based analysis identifies four major categories of failure: grounding errors, competency gaps, knowledge deficiencies, and redundant loops. For each category, the LLM proposes a corresponding solution: visual search, terminal execution, knowledge support, and repetition warnings. A human then selects a target solution, minimally adjusts the patch code generated by the LLM, and integrates it into the agent's workflow, which is then re-evaluated on OSWorld (see Appendix~\ref{sec:human}). The enhanced agent reaches 48.9\% accuracy—an absolute improvement of +6.6 points at no additional training cost and only modest inference overhead. This gain directly expands the proportion of usable environments that can contribute high-quality trajectories to the successful-case loop. A further generalization study shows consistent improvements across different graphical user interface (GUI) benchmarks, indicating that our identified failure categories capture general patterns of GUI-agent deficiencies. Moreover, we observe consistent gains across agents of different families and scales, with stronger models benefiting the most—highlighting the potential of our framework to further amplify the performance of increasingly capable computer-use agents.

Our main contribution is three-fold: (1) we identify a fundamental inefficiency in current agent-improvement pipelines, namely the under-utilization of failed trajectories in success-driven learning loops; (2) we propose a complementary failure-driven self-improvement loop, a data-centric paradigm that leverages failed trajectories to derive inference-time behavioral improvements through LLM-based analysis and lightweight human verification; and (3) we provide, to our knowledge, the first empirical evidence that such a loop can meaningfully improve computer-use agents, demonstrating a 6.6-point absolute gain on OSWorld. Together, our results suggest that failed trajectories are not merely noise but a structured source of supervision, and that incorporating them into an LLM-guided improvement process offers a practical path toward more efficient and continually improving computer-use agents.

\section{Related Work}

\paragraph{Computer-use agents.} Computer-use agents, driven by multimodal large language models (MLLMs), automate graphical user interface (GUI) interactions and real-world tasks~\cite{wang2025opencuaopenfoundationscomputeruse,qin2025uitarspioneeringautomatedgui,agashe2024agent,agashe2025agent,yuan2025surfer2,song2025coact1,feizi2025grounding,yao2025agentic,hu2025osagents}. These agents process instructions and visual states (e.g., screenshots) to generate actions such as mouse clicks, keyboard inputs, or commands, enabling navigation in computer environments. Early systems such as Agent S~\cite{agashe2024agent} and its successor Agent S2~\cite{agashe2025agent} leverage powerful MLLMs as planners and controllers, introducing hierarchical planning, grounding specialists, and task decomposition to handle long-horizon GUI tasks. More recent work develops native agents for efficient multi-step GUI interaction, such as UI-TARS~\cite{qin2025uitarspioneeringautomatedgui}, OpenCUA~\cite{wang2025opencuaopenfoundationscomputeruse}, and Surfer 2~\cite{yuan2025surfer2}, which use a single agent to perform both planning and control. Benchmarks like OSWorld~\cite{xie2024osworld} offer verifiable, open-ended tasks for evaluating performance in areas including file management, web browsing, and software operation. In this work, we leverage the state-of-the-art open-source model OpenCUA~\cite{wang2025opencuaopenfoundationscomputeruse} to explore inference-time self-improvement of a computer-use agent on the widely used OSWorld benchmark~\cite{xie2024osworld}.

\paragraph{Self-improvement of computer-use agents.} Self-improvement enables AI agents to iteratively refine their behavior while minimizing dependence on human annotations~\cite{juan2025surveyselfevolving}. Most existing self-improvement strategies rely on success-driven self-training loops. In frameworks such as OpenCUA, UI-TARS, and Mobile-Agent~\cite{wang2025opencuaopenfoundationscomputeruse,qin2025uitarspioneeringautomatedgui,ye2025mobileagentv3fundamentalagentsgui}, the agent executes tasks, retains only successful trajectories, and periodically fine-tunes on this curated dataset. This paradigm reduces the need for human demonstrations but discards the majority of experience—namely, failed trajectories—even though they contain valuable diagnostic signals. Alternative approaches include self-evolving agents such as SEAgent~\cite{sun2025seagent}, SEA~\cite{huo2025sea}, and UI-Genie~\cite{zhou2025uigenie}, which learn from experience through trial-and-error or iterative boosting. Other methods employ self-evolutionary reinforcement learning for visual grounding~\cite{yuan2025enhancing} or efficient training with reduced data reliance~\cite{he2025efficient}. Our work introduces a complementary failure-driven loop. Instead of relying on additional training, we use an LLM to analyze failed trajectories at inference time, extract systematic failure modes, and apply targeted code or strategy patches that immediately modify the agent's behavior. This yields cost-effective enhancements that seamlessly complement existing success-driven self-training loops.

\paragraph{Error analysis of computer-use agents.} Although recent computer-use agents have made promising progress toward general-purpose computer use, several fundamental bottlenecks remain~\cite{zhang2025large,hu-etal-2025-os}. GUI environments are visually and structurally complex, containing text, icons, widgets, and hierarchical layouts that vary across applications. Prior work highlights the difficulty of building reliable multimodal representations for screens~\cite{wang2021screen2wordsautomaticmobileui, Wu_2021}, while recent GUI-focused grounding studies show that incorrect or incomplete perception is a leading failure mode~\cite{zheng2024gpt4visiongeneralistwebagent,liu2025infigui,wu2025see}. Moreover, many GUI tasks demand multi-step workflows and contingent decision-making over evolving interface states; recent web and desktop benchmarks show that reasoning errors grow with trajectory length and that failure recovery remains limited~\cite{deng2023mind2webgeneralistagentweb, zhou2024webarenarealisticwebenvironment, xie2024osworldbenchmarkingmultimodalagents}. In this work, we use an LLM to analyze failed trajectories of computer-use agents and identify representative failure modes, producing findings comparable to—and in some cases beyond—those of previous studies. We further use the LLM to propose inference-time corrections for these errors.

\section{Method}
\label{sec:method}
\begin{figure}[t] 
  \centering
  \includegraphics[width=\linewidth]{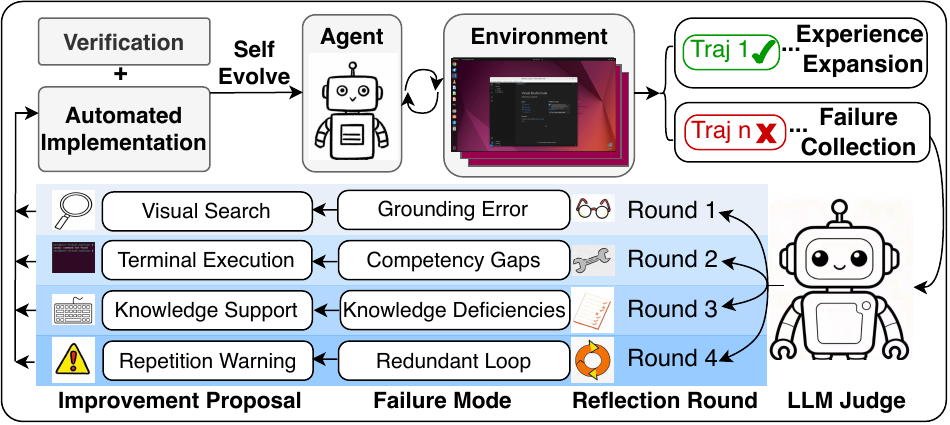} 
  \caption{Overview of our failure-case loop, a self-evolving framework. In each round, failed trajectories are collected through agent rollout; a large language model (LLM) then acts as a meta-controller that diagnoses failure modes, proposes inference-time solutions, and applies code patches. The recipe for OSWorld is generated over multiple rounds of this failure-case loop.}
  \label{fig:gui_main}
\end{figure}
In this section, we elaborate on the \textbf{failure-case loop}, a data-centric self-improvement mechanism that leverages failed trajectories collected during task execution, categorizes systematic failure modes, and synthesizes targeted corrective strategies to address them. The framework organizes failed trajectories into structured, reusable behavioral improvements, enabling the agent to iteratively refine its execution at inference time. We then present a systematic analysis of the prevalent failure patterns observed in GUI-agent interactions, together with a principled and generalizable solution for each.

\begin{table}[tb]
\caption{Ablation study of the proposed inference-time optimization strategies. Each strategy improves the inference-time performance of the OpenCUA-72B model on the OSWorld small set, and combining all strategies further boosts performance to 52.74\%.}
\centering
\begin{tabular}{l c}
\toprule
\textbf{Method} & \textbf{Performance} \\
\midrule
\textbf{Baseline:} \\OpenCUA-72B (30 steps) & 41.67 \\
\midrule
+ Visual Search & 47.22 \\
+ Repetition Detection & 44.40 \\
+ Terminal Execution & 47.19 \\
+ Knowledge Support & 44.44 \\
\midrule
\textbf{+ Full Method (Ours)} & \textbf{52.74} \\
\bottomrule
\end{tabular}
\label{tab:self_evolving_ablation}
\end{table}

\subsection{Framework Overview}
We introduce a self-evolving framework, illustrated in Figure~\ref{fig:gui_main}, that systematically leverages failed trajectories observed during execution. The LLM-guided loop identifies and categorizes failure modes and synthesizes targeted improvement strategies to mitigate them. Over time, these strategies accumulate into a continually updated \emph{set of optimization strategies}, enabling the agent to refine its execution in a data-driven manner. This cycle is computationally efficient and progressively enhances the agent's inference-time capabilities by making effective use of large numbers of failed trajectories.
\paragraph{Failure Experience Collection.}
Within the given verifiable environment, the agent performs rollouts to generate a diverse set of execution trajectories. Each trajectory is evaluated by the environment's built-in reward function, which provides quantitative feedback reflecting task-specific performance. Failed trajectories are preserved as the basis for subsequent diagnosis and improvement.

\paragraph{LLM-Guided Self-Improvement Loop.}
To extract experience from failed trajectories, we invoke a large language model to conduct targeted error analysis. Each problematic trajectory is fed into the LLM together with its \emph{instruction}, \emph{action history}, and \emph{thought process}, allowing the model to reason about failure modes in context. The LLM then performs a structured diagnosis, identifying underlying bottlenecks such as grounding deficiencies, action redundancy, and operational competency gaps. Based on this analysis, it proposes inference-time improvements, including conceptual adjustments and direct \texttt{code implementations} that are \emph{lightly verified by humans}, to enhance the agent's behavior. The LLM introduces four core strategies iteratively, selecting one at each round according to the dominant failure mode, and these strategies accumulate across rounds. After applying an adjustment, the agent executes new rollouts, producing updated failure cases that are fed back into the same diagnostic loop, thereby enabling a continuous refinement cycle. A detailed explanation is provided in Appendices~\ref{appendix:selfimprovementloop}, \ref{appendix:fullalgorithm}, and \ref{appendix:example}.

Through several rounds of the improvement loop, our framework identifies a set of recurring failure patterns that hinder the agent's progress in complex multimodal environments. These deficiencies reveal fundamental limitations in grounding, control, knowledge, and procedural efficiency.

\subsection{Grounding Errors}

\begin{figure}[tb]
  \centering
  \begin{subfigure}[t]{0.49\linewidth}
    \centering
    \includegraphics[width=\linewidth]{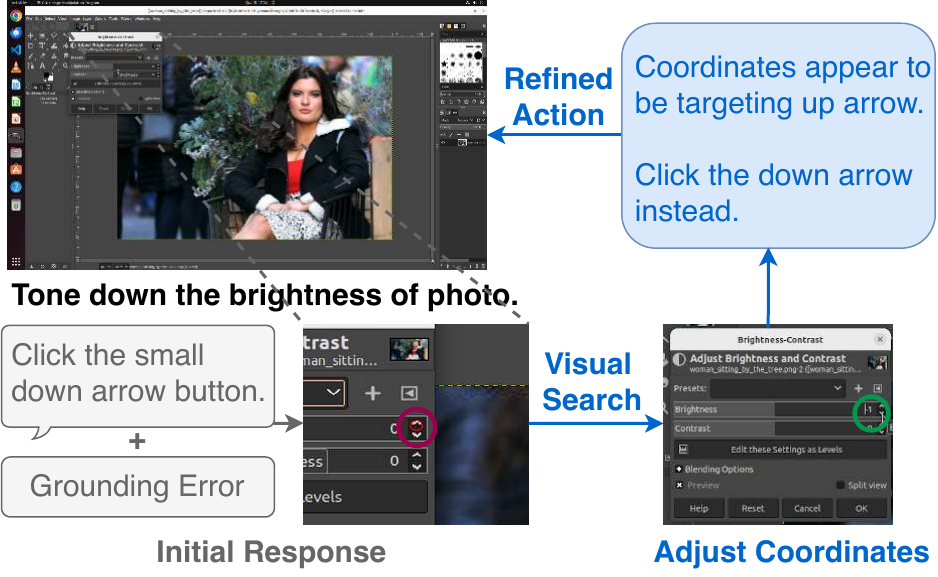}
    \caption{Visual search strategy. The agent initially mis-localizes the down-arrow in the settings panel; once it is offered a zoomed-in view, it correctly refines its grounding.}
    \label{fig:visual_search_case}
  \end{subfigure}
  \hfill
  \begin{subfigure}[t]{0.49\linewidth}
    \centering
    \includegraphics[width=\linewidth]{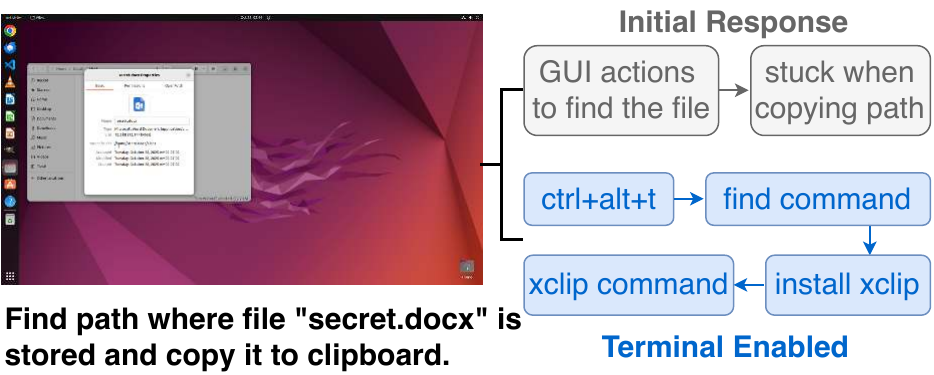}
    \caption{Terminal execution strategy. Whereas searching for a file through GUI actions is complicated, three command executions complete the task successfully.}
    \label{fig:terminal_use_case}
  \end{subfigure}

  \caption{Case studies of the visual search and terminal execution strategies. In both cases, our framework guides the agent to correctly improve its behavior.}
  \label{fig:visual_search_terminal_use}
\end{figure}

\paragraph{Failure Mode Analysis.}
In the first few rounds of LLM-driven failure analysis, grounding errors emerge as one of the most common bottlenecks for GUI agents. Understanding interface layouts, locating functional elements, and estimating visual coordinates remain particularly challenging, especially in high-resolution or visually cluttered settings. As a result, the agent often misinterprets visual cues or fails to align its actions with the intended targets, leading to inaccurate operations and suboptimal task outcomes.
\paragraph{Proposed Strategy: Visual Search.}

To improve grounding accuracy, we introduce a \textbf{visual search mechanism} that enables multi-round localization and self-refinement for grounding-related actions. Unlike prior GUI grounding approaches that primarily emphasize perception or pre-action grounding, our method integrates visual search as a \emph{first-class module for post-action visual self-verification}.

Specifically, whenever the agent executes a spatially grounded operation such as \texttt{click}, \texttt{moveto}, or \texttt{dragto}, the framework extracts a fixed-size region centered at the target location and generates a zoomed-in view to facilitate fine-grained visual interpretation and spatial reasoning. This localized view is augmented with a visual hint, a red circular marker indicating the original action position. The agent is then prompted to verify its action conditioned on the cropped screenshot, task instruction, action history, and ongoing reasoning context, and to revise the grounding decision if inconsistencies are detected. A more detailed explanation is provided in Appendix~\ref{appendix:visualsearch}.

By explicitly closing the loop between action execution and visual verification, this interactive refinement process substantially improves the precision and reliability of grounding in high-resolution, cluttered graphical user interfaces.
\paragraph{Outcome.}
On the OSWorld small set, visual search improves performance from 41.67 to 47.22, as shown in Table~\ref{tab:self_evolving_ablation}. Consider the case in Figure~\ref{fig:visual_search_case}: the agent is asked to tone down the brightness of a photo in GIMP. When it first triggers a grounding error (clicking the up-arrow while attempting to lower the brightness), our framework provides a zoomed-in view together with a visual prompt. The agent then reflects on its action and proposes a corrected coordinate, leading to task success.

\subsection{Competency Gaps}

\paragraph{Failure Mode Analysis.}
Another recurring failure mode identified by the LLM is the agent's competency gaps. Although operating systems provide reliable, high-level control interfaces—most notably terminals, which allow direct execution of precise commands—current GUI agents rarely leverage these capabilities, defaulting instead to low-level mouse- or cursor-based manipulation. This reliance on fine-grained interface interaction increases execution volatility and compounds error accumulation, hindering agent performance.

\paragraph{Proposed Strategy: Terminal Execution.}
Accordingly, the LLM proposes terminal use, activated through a dedicated hotkey, to perform system-level operations. This capability allows the agent to translate operation-intensive tasks into concise command-line instructions, bypassing complex graphical interactions and accomplishing many workflows with greater accuracy and efficiency. A more detailed explanation of this strategy is provided in Appendix~\ref{appendix:terminalexecution}.

\paragraph{Outcome.}
In the case shown in Figure~\ref{fig:terminal_use_case}, the agent is asked to locate a specific file and copy its path to the clipboard. Terminal execution provides a direct way to complete this task; otherwise the agent would attempt it through GUI actions and get stuck when copying the path. This strategy improves performance on the OSWorld small set by 5.52 points.

\subsection{Knowledge Deficiencies}

\begin{figure}[tb]
  \centering
  \begin{subfigure}[t]{0.49\linewidth}
    \centering
    \includegraphics[width=\linewidth]{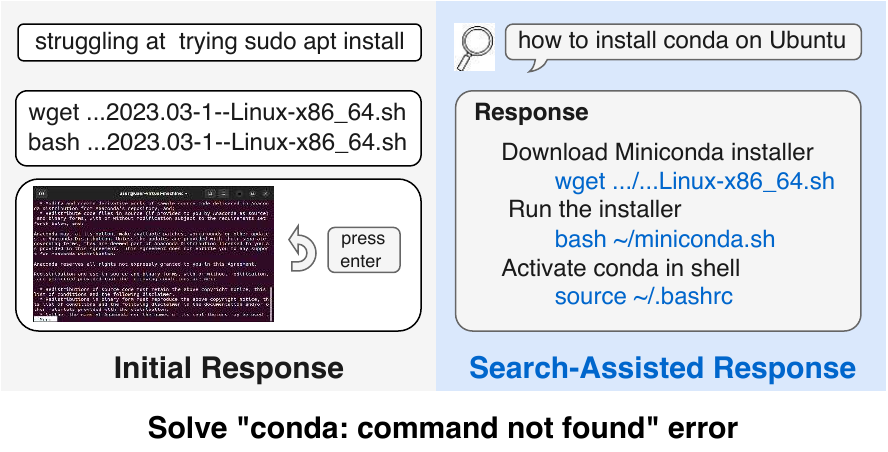}
    \caption{Knowledge support via search engine. The agent is unfamiliar with the commands to install conda, but querying GPT-5-mini helps it solve the task.}
    \label{fig:search_engine_case}
  \end{subfigure}
  \hfill
  \begin{subfigure}[t]{0.49\linewidth}
    \centering
    \includegraphics[width=\linewidth]{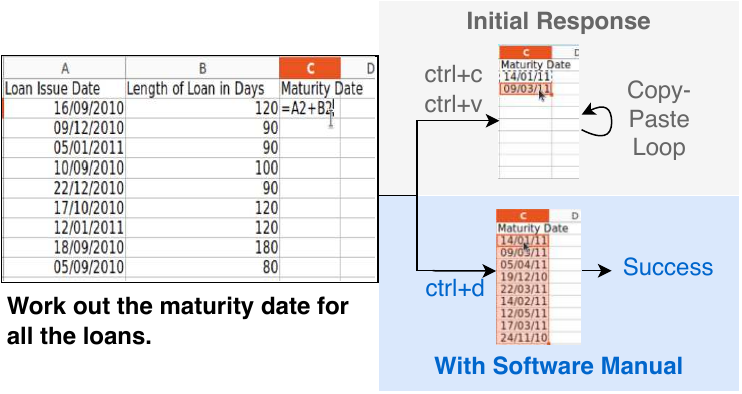}
    \caption{Knowledge support via software manual. Given LibreOffice hotkeys, the agent bypasses a complex copy--paste procedure and fills all cells with a single hotkey.}
    \label{fig:hotkey_case}
  \end{subfigure}

  \caption{Case studies of the knowledge support strategy. In both cases, our framework guides the agent to correctly improve its behavior.}
  \label{fig:visual_knowledge}
\end{figure}
\paragraph{Failure Mode Analysis.}
During agent execution and iterative evolution, the LLM identifies gaps in the agent's initial knowledge as a major source of failures, manifesting in patterns such as insufficient multi-step planning and repeated action loops. These issues arise when task-relevant information falls outside the agent's internal context, particularly for domain-specific operations such as application-level procedures or command-line syntax. Lacking access to external references, the agent cannot reliably execute tasks that depend on specialized or factual knowledge.

\paragraph{Proposed Strategy: Knowledge Support.}
To mitigate knowledge deficiencies, we implement two complementary mechanisms: a \textbf{software-manual retriever} that accesses a curated repository of application manuals and hotkey references (Appendix~\ref{appendix:softwaremanual}), and a \textbf{search-engine interface} that fetches relevant information beyond the model's internal context window (Appendix~\ref{appendix:searchengine}). Together, these components enable the agent to acquire software-level operational knowledge and to autonomously query external knowledge bases when uncertainty arises, improving its capacity to execute domain-specific and knowledge-intensive tasks.

\paragraph{Outcome.}
Knowledge support improves the performance of the OpenCUA-72B agent from 41.67 to 44.44. An example appears in Figure~\ref{fig:search_engine_case}, where the agent is asked to resolve a ``conda: command not found'' error. Although the agent initially does not know how to do this, by retrieving a result from GPT-5-mini it learns the correct commands to install conda.

A second example is the LibreOffice case in Figure~\ref{fig:hotkey_case}, where the agent must compute a maturity date for every row of data. While it initially struggles with a \texttt{Ctrl+C}/\texttt{Ctrl+V} loop, after being given the software manual it learns to use \texttt{Ctrl+D} to fill the selected range instantly, completing the task efficiently.

\subsection{Redundant Loops}
\paragraph{Failure Mode Analysis.}
The most common failure mode identified by the LLM is redundant action loops, which usually arise when the agent fails to recognize that it is stuck in repetitive or ineffective behavior. For example, after an operation fails, the agent may repeatedly execute the same action sequence without reflecting on the underlying cause or considering alternatives, effectively falling into an action loop. In other cases, the agent performs a sequence of clicks or UI interactions but does not verify whether the expected state transition has actually occurred. As a result, it continues its current plan even when the interface remains unchanged, leading to prolonged stagnation and inefficient exploration. This lack of self-awareness not only wastes many inference steps but also prevents timely recovery from task deviations, ultimately reducing task-completion reliability and overall success rates.

\begin{figure}[tb]
  \centering
  \includegraphics[width=0.6\linewidth]{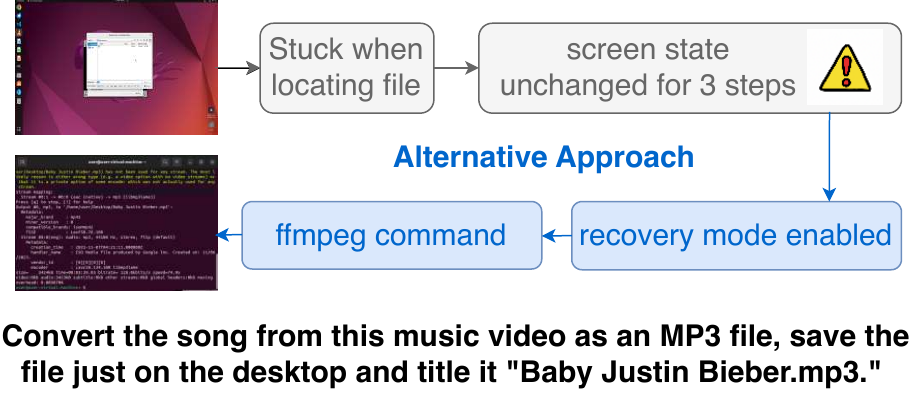}
  \caption{Repetition-warning strategy. The agent initially gets stuck at a screen state; after three rounds of unproductive attempts, recovery mode is triggered, prompting the agent to adopt an alternative strategy and succeed.}
  \label{fig:repetition_detection}
\end{figure}

\paragraph{Proposed Strategy: Repetition Warnings.}
To mitigate redundant action loops, the LLM introduces a systematic \textbf{repetition-detection mechanism} that monitors recent interaction history within a fixed-size sliding window of the agent's \emph{thought}, \emph{action}, and \emph{screen-state} traces. Rather than requiring strictly consecutive repetitions, the mechanism examines the \emph{frequency and recurrence patterns} within the window, flagging repetition once occurrences exceed a predefined threshold, which signals stagnation. Specifically:

\begin{itemize}
    \item \textbf{Thought repetition.} We detect repetition when the same or semantically equivalent low-level planning statements reappear multiple times within the window, indicating that the agent's reasoning loop is not progressing toward a new decision.
    \item \textbf{Action repetition.} If identical action commands (e.g., the same \texttt{pyautogui} interaction) occur frequently across the window, we classify this as action repetition, suggesting that the agent is executing without adapting to task feedback.
    \item \textbf{Screen-state repetition.} We encode each screen state into a compressed representation capturing structural and textual cues. If the encoded states remain unchanged across the window despite multiple actions, we detect screen-state repetition, signaling that the agent's actions are not producing the intended interface progress.
\end{itemize}

When any of these patterns is detected, the mechanism issues an explicit warning to the agent and activates a \textbf{recovery mode}. In this mode, the agent is encouraged to reformulate its plan or switch to an alternative interaction strategy, such as exploring a different UI pathway or invoking an auxiliary tool (e.g., a search or system-level command interface). This intervention prevents prolonged stagnation, reduces wasted inference steps, and substantially improves task-completion reliability. A more detailed description of the repetition rule and intervention is provided in Appendix~\ref{appendix:repetitionwarning}.

\paragraph{Outcome.}
Table~\ref{tab:self_evolving_ablation} shows the agent's performance before and after the repetition-detection mechanism is applied; it improves performance by 2.73 points. An example is shown in Figure~\ref{fig:repetition_detection}: the agent normally gets stuck halfway, but after detecting that the screen state remains unchanged for 3 rounds, it switches to a command-line approach to finish the task.

\subsection{Overall Effectiveness}
\label{sec:ablation}
\begin{figure}[tb] 
  \centering
  \includegraphics[width=\linewidth]{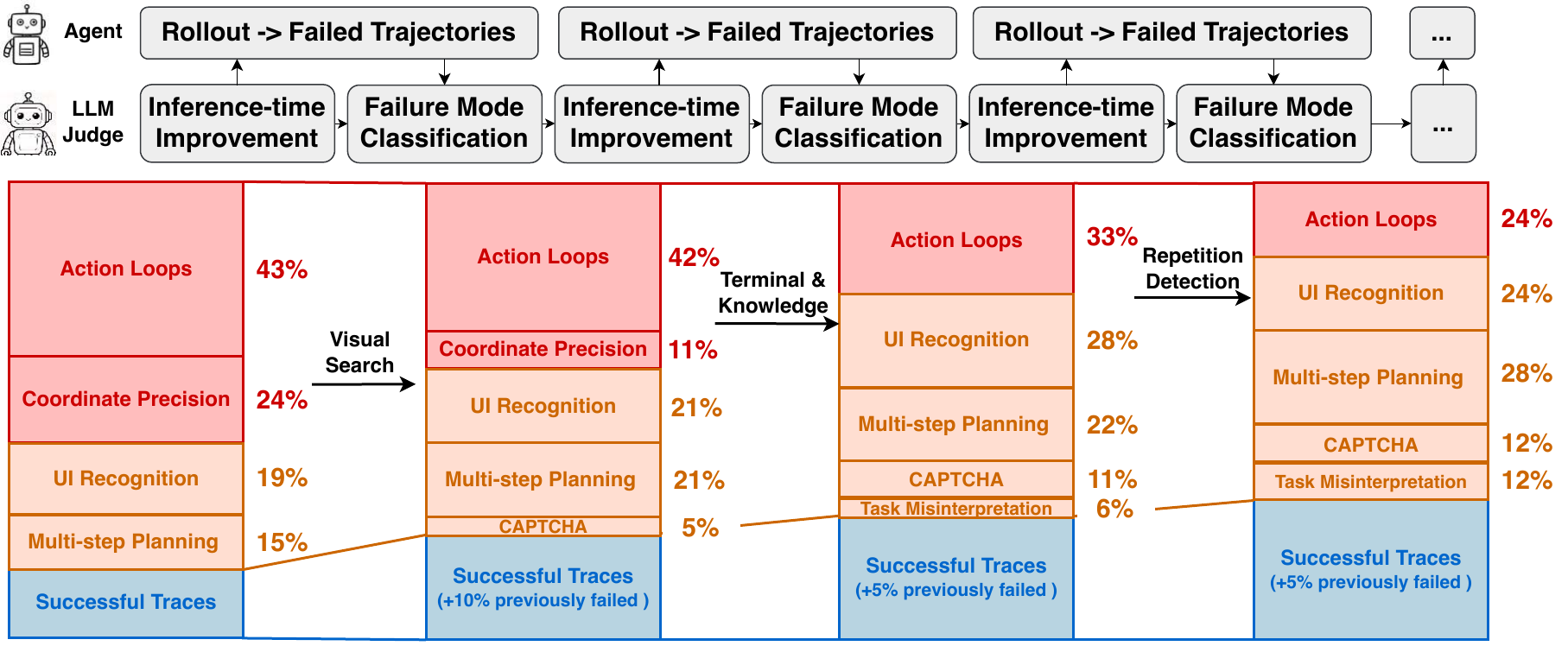} 
  \caption{Failure-mode distribution. The agent's initial failures mainly consist of grounding errors, redundant action loops, and a lack of recovery; our self-evolving framework helps the agent overcome these deficiencies. The remaining failure modes span task comprehension, reasoning consistency, and visual understanding, reflecting higher-level cognitive aspects of agent ability. (Failure-mode proportions are approximately proportional to bar heights; the ``success'' portion is illustrative only and not to scale.)}
  \label{fig:failure_mode}
\end{figure}
\paragraph{Outcome.}
Table~\ref{tab:self_evolving_ablation} reports an ablation study of each improvement strategy introduced above, as well as the overall effect when all of them are combined. The OpenCUA-72B model is evaluated on the OSWorld small set—a lightweight subset of OSWorld used for rapid iteration—with a 30-step task limit, in contrast to the full-set, 100-step evaluation reported in Table~\ref{tab:osworld_comparison}.

Integrating all of these techniques into a unified framework yields a substantial performance boost on the OSWorld small set—one that clearly exceeds what any single technique achieves on its own. Each component addresses a different aspect of the agent's limitations, and their combined use enables the system to operate with greater robustness, adaptability, and efficiency. Overall, the OpenCUA-72B score increases from 41.67 to 52.74, highlighting both the contribution of each individual technique and the benefit of combining their complementary strengths.

\paragraph{Failure Mode Transition.}
As summarized in Figure~\ref{fig:failure_mode}, we observe a clear shift in the distribution of failure modes, as categorized by the LLM, before and after applying our framework. Because a single failed trajectory may exhibit more than one failure mode, these proportions are not mutually exclusive and need not sum to 100\%. In the earlier version, failures were dominated by action loops (43\%) and coordinate-precision errors (24\%), indicating that the agent frequently struggled with precise grounding, knowledge deficiency, and self-recovery. UI-recognition errors (19\%) and insufficient multi-step planning (15\%) further reflected limitations in perceptual grounding and multi-step reasoning.

After incorporating our improvements, these issues are largely mitigated: the proportion of failed trajectories declines, and the remaining failures shift toward higher-level cognitive aspects of agent ability. After four rounds of our failure-case loop, the remaining failure modes consist of task misinterpretation (12\%), CAPTCHA failures (12\%), UI-recognition errors (24\%), insufficient multi-step planning (28\%), and action loops (24\%).

This transition demonstrates that our framework effectively resolves the majority of low-level bottlenecks, such as repetitive action loops, grounding errors, and the lack of a self-recovery mechanism. The remaining challenges shift toward higher-level cognitive aspects—task comprehension, reasoning consistency, and visual understanding—revealing deeper limitations in the model's semantic and reasoning capabilities.

\section{Results}
\label{sec:results}

This section evaluates our failure-case loop on realistic computer-use tasks. In short, it raises OpenCUA-72B from 42.3 to 48.9 on OSWorld (+6.6 points, +15.6\% relative) at no additional training cost, our ablation shows each strategy contributes and combining them works best, and the same loop transfers across model scales and four heterogeneous benchmarks without retraining. We detail the setup, main results, ablations, and generalization studies below.

\subsection{Experimental Setup}

We design a comprehensive evaluation to test whether our \emph{failure-case loop} enhances agent performance in realistic operating environments. The experiments are conducted on a complex computer-use benchmark, with systematic baseline comparisons, detailed ablation analysis, and generalization tests.

\paragraph{Benchmark.}
We validate our framework on the \textbf{OSWorld}~\cite{xie2024osworld} benchmark, chosen for its real-world complexity and broad task coverage across diverse operating-system scenarios such as document editing and file management. This environment provides a high-fidelity testbed for evaluating both the agent's \emph{grounding accuracy} and \emph{operational competence} under realistic conditions.

\paragraph{Baseline.}
As the reference model, we employ the open-source \textbf{OpenCUA-72B}~\cite{wang2025opencuaopenfoundationscomputeruse}, which has strong general reasoning capabilities and partial familiarity with OSWorld's action space. Its balanced performance across general and system-level tasks makes it a suitable baseline for evaluating the improvements brought by our failure-driven self-improvement framework.

\paragraph{Failure-Case Loop.}
In each round, failed-trajectory collection, trajectory analysis, failure diagnosis, and code refinement are conducted by \textbf{Claude 4.5 Sonnet}, which serves as a meta-level reasoner. It analyzes erroneous trajectories, identifies behavioral bottlenecks, and proposes targeted improvements that are incorporated into the agent's inference-time optimization loop with light human verification. In practice, the human role is limited to lightweight verification and candidate selection: over 97\% of LLM-generated refinements are accepted without modification, indicating that the loop is predominantly model-driven. We run the loop for four rounds, identifying failure modes and proposing inference-time solutions as discussed in Section~\ref{sec:method}.

\paragraph{Meta-Controller.}
We select the meta-controller from among Claude 4.5 Sonnet, GPT-5.2, Gemini 3 Flash, and the open-source Qwen3-VL-32B-Instruct by comparing their performance. While GPT-5.2 and Qwen3-VL-32B-Instruct identify certain failure modes, their analyses are less comprehensive and less consistent in proposing implementable refinements, and Gemini 3 Flash exhibits comparatively weaker code-level implementation ability. Since the meta-controller must jointly analyze failures, propose actionable improvements, and implement code modifications, Claude 4.5 Sonnet demonstrates the strongest overall performance and is therefore adopted in our framework.

\paragraph{Evaluation and Ablation.}
We evaluate the refined framework on OSWorld and compare it against both the baseline and other high-performing systems on the leaderboard. To ensure a fair comparison, our framework retains exactly the same tool access, environment configurations, and action spaces as the baseline agents. To quantify the contribution of each component, Section~\ref{sec:ablation} reports ablations on the OSWorld small set that isolate \textbf{visual search}, \textbf{terminal execution}, \textbf{knowledge support}, and \textbf{repetition warnings}, measuring their individual effects on \textbf{task success rate}. These analyses show that the integrated system outperforms any individual module, combining their complementary strengths into a single, more robust agent.

\paragraph{Cross-Model Generalization.}
To assess the breadth and robustness of our method's generalization, we apply the same failure-driven self-improvement loop to models of different capacities and origins—most notably \textbf{OpenCUA-32B}~\cite{wang2025opencuaopenfoundationscomputeruse} and \textbf{GUI-Owl-32B}~\cite{ye2025mobileagentv3fundamentalagentsgui}. These two models differ substantially in architecture, training philosophy, and pretraining sources, providing a meaningful testbed for the transferability of our framework across heterogeneous backbones. We report their performance on OSWorld with the improvement strategies of Section~\ref{sec:method} applied.

\paragraph{Cross-Benchmark Generalization.}
To evaluate cross-benchmark generalization, we directly transfer the failure-derived patches mined from OSWorld to heterogeneous benchmarks: \textbf{OmniACT}~\cite{kapoor2024omniactdatasetbenchmarkenabling} for desktop-level agent tasks, \textbf{AndroidControl}~\cite{li2024effectsdatascaleui} for mobile environments, \textbf{ScreenSpotPro}~\cite{li2025screenspotproguigroundingprofessional} for advanced GUI grounding, and \textbf{WebVoyager}~\cite{he2024webvoyagerbuildingendtoendweb} for web-based multimodal interaction. We use \textbf{Qwen3-VL-32B-Instruct} as the default backbone for these tests under identical evaluation protocols.

\subsection{Results}
\begin{table}[tb]
\caption{Comparison with state-of-the-art methods on the OSWorld~\cite{xie2024osworld} benchmark. The second column reports the number of steps used by each agent, where one step is a single agent--environment interaction (action + observation), and success rate (\%) is the evaluation metric. For our framework, the main results are the mean $\pm$ standard deviation across 3 independent runs; $^{*}$ denotes configurations evaluated in a single run due to resource constraints.}
\vspace{-1em}
\small
    \centering
    \begin{tabular}{lcc}
        \toprule
        \textbf{Agent Model} & \textbf{Step} & \textbf{Success Rate}\\
        \midrule
        \multicolumn{3}{l}{\textit{Proprietary Models}} \\
        Claude 3.7 Sonnet~\cite{claude37} & 100 & 28.0 \\
        OpenAI CUA 4o~\cite{openAI_o3_o4_mini} & 200 & 38.1 \\
        UI-TARS-1.5~\cite{qin2025ui} & 100 & 42.5   \\
        OpenAI CUA o3~\cite{openAI_o3_o4_mini} & 200 & 42.9   \\
        \midrule
        \multicolumn{3}{l}{\textit{Open-Source Models}} \\
        Aria-UI w/ GPT-4o~\cite{yang2024aria} & 15 & 15.2 \\
        Aguvis-72B w/ GPT-4o~\cite{xu2024aguvis} & 15 & 17.0  \\
        UI-TARS-72B-SFT~\cite{qin2025ui} & 50 & 18.8 \\
        Agent S w/ Claude-3.5-Sonnet ~\cite{agashe2024agent}  & 15 & 20.5 \\
        Agent S w/ GPT-4o~\cite{agashe2024agent}  & 15 & 20.6   \\
        UI-TARS-72B-DPO~\cite{qin2025ui} & 15 & 22.7   \\
        UI-TARS-72B-DPO~\cite{qin2025ui} & 50 & 24.6 \\
        UI-TARS-1.5-7B ~\cite{qin2025ui} & 100 & 26.9  \\
        Jedi-7B w/ GPT-4o~\cite{xie2025scaling} & 100 & 27.0  \\
        Agent S2 w/ Claude-3.7-Sonnet~\cite{agashe2025agent} & 50 & 34.5 \\
        Agent S2 w/ Gemini-2.5-Pro~\cite{agashe2025agent} & 50 & 41.4 \\
        GUI-Owl-32B~\cite{ye2025mobileagentv3fundamentalagentsgui}  & 100 & 19.0$^{*}$\\
        \quad + Ours & 100 & \textbf{21.3}$^{*}$\\
        OpenCUA-32B~\cite{wang2025opencuaopenfoundationscomputeruse} &  100 & 34.5$^{*}$ \\
        \quad + Ours & 100 & \textbf{38.2}$^{*}$\\
        OpenCUA-72B~\cite{wang2025opencuaopenfoundationscomputeruse} &  100 & $42.3 \pm 2.6$ \\
        \quad + \textbf{Ours (Main)} & 100 & $\mathbf{48.9 \pm 1.2}$ \\
        
        \bottomrule
    \end{tabular}
    \label{tab:osworld_comparison}
\end{table}
\begin{table}[t]
\caption{Cross-benchmark generalization results. The results show that our method captures transferable failure patterns and outperforms the base agents. All experiments are repeated three times except for WebVoyager, and we report the mean and standard deviation.}
\label{tab:generalization}
\centering
\small
\setlength{\tabcolsep}{4pt}
\begin{tabular}{lccccc}
\toprule
 & OmniACT & AndroidControl & ScreenSpotPro & WebVoyager \\
\midrule
Base  & 4.77$\pm$0.02 & 28.37$\pm$0.13 & 27.50$\pm$0.35 & 23.80 \\
Ours  & \textbf{6.90$\pm$0.10} & \textbf{36.23$\pm$0.22} & \textbf{30.74$\pm$0.27} & \textbf{27.90} \\
\bottomrule
\end{tabular}
\vspace{-3mm}
\end{table}
\paragraph{Main Results.}
Our primary evaluation uses the \textbf{OpenCUA-72B} model as the backbone for our failure-case loop, with performance measured on the \textbf{OSWorld} benchmark over 100 steps.

As shown in Table~\ref{tab:osworld_comparison}, applying our framework improves performance from \textbf{42.3} to \textbf{48.9}, an absolute gain of \textbf{+6.6} points (\textbf{+15.6\%} relative). This highlights the synergy between a strong base model and our failure-driven self-improvement mechanism, which progressively refines reasoning, grounding, and recovery throughout interaction.

Importantly, this improvement is achieved without any additional training cost, since all enhancements are applied at inference time. The system introduces only modest computational overhead—about an 8\% increase in runtime—while reducing interaction steps by roughly 15\%, indicating improved execution efficiency.

\paragraph{Generalization Across Models, Scales, and Environments.}
To verify the generalization of our approach, we apply the same failure-driven self-improvement framework to models of different capacities and origins, specifically \textbf{OpenCUA-32B} and \textbf{GUI-Owl-32B}. As reported in Table~\ref{tab:osworld_comparison}, across both settings we observe consistent \textbf{+10–12\%} relative gains regardless of model size or provenance. These uniform improvements indicate that the proposed recipe generalizes reliably across scales and sources, reinforcing the view that our framework acts as a scalable, model-agnostic amplifier of base-model capabilities.

For cross-benchmark generalization, as shown in Table~\ref{tab:generalization}, incorporating failure-derived patches mined from OSWorld yields consistent improvements across all evaluated benchmarks. The largest gains appear on WebVoyager (+4.10 points) and AndroidControl (+7.86 points), both of which involve long-horizon planning and interactive command execution. This further supports the hypothesis that failure trajectories mined from OSWorld capture generalizable error patterns rather than environment-specific heuristics.

\section{Discussion}

This work presents a self-evolving framework centered on a structured \textbf{failure-case loop} for GUI agents, which leverages failed trajectories at inference time instead of discarding them. Rather than treating errors as terminal outcomes, the framework organizes them into actionable signals—identifying bottlenecks in grounding, control, knowledge usage, and procedural efficiency—and applies targeted inference-time strategies to mitigate them in a semi-automatic manner.

The resulting plug-and-play recipe operates entirely at inference time, requiring no additional training or data collection, yet consistently strengthens agent performance on OSWorld, a highly challenging computer-use environment. Applied to a strong baseline such as \textbf{OpenCUA-72B}, the framework improves performance from \textbf{42.3} to \textbf{48.9}, with similarly robust \textbf{+10–12\%} relative gains across models of different origins and scales, including \textbf{OpenCUA-32B} and \textbf{GUI-Owl-32B}. It also generalizes well to other GUI-agent benchmarks.

This failure-case loop complements existing success-based loops, jointly improving trajectory quality by amplifying successful behaviors and correcting failures. Importantly, we show that substantial improvements can be achieved at inference time through structured interventions, without any weight updates.

The framework further scales through a sequential, failure-driven update process in which each iteration targets a newly identified failure mode rather than accumulating ad-hoc patches. As shown in our experiments, stacking four structured updates consistently outperforms any single update, indicating that improvements compose constructively and scale reliably.

Moreover, learned reward models trained on prior interaction data could be integrated into our framework to automatically identify erroneous or suboptimal trajectories without human feedback, further improving scalability and enabling continuous self-refinement in open-ended environments.

Overall, these results demonstrate a \textbf{scalable, general, and data-efficient} route for advancing GUI agents. By systematically leveraging failure cases as continual inference-time signals, the framework enables agents to extract substantially more value from their interactions with the environment, paving the way toward more efficient and self-improving systems.

\paragraph{Acknowledgments.} We gratefully acknowledge Lambda, Inc. for providing partial computational support for this project. S.Y. is a Chan Zuckerberg Biohub — San Francisco Investigator.

\newpage

{
    \small
    \bibliographystyle{splncs04}
    \bibliography{main}
}

\appendix
\clearpage

\setcounter{page}{1}

\title{Supplementary Materials for \\ Learning from Failure: Inference-Time Self-Improvement for Computer-Use Agents}

\titlerunning{Inference-Time Self-Improvement for Computer-Use Agents}
\author{
Xueqiao Sun\inst{1,2} \and
Xiaohan Wang\inst{1} \and
Ludwig Schmidt\inst{1} \and
Serena Yeung-Levy\inst{1,\dagger} \and
Yuhui Zhang\inst{1,\dagger} 
}

\authorrunning{Sun et al.}

\institute{$^1$Stanford University, Stanford, CA 94305, USA \\ $^2$Tsinghua University, Beijing, 100084, China}

{
  \renewcommand{\thefootnote}{\fnsymbol{footnote}}
  \footnotetext[0]{$^\dagger$Corresponding authors: \texttt{\{yuhuiz,syyeung\}@stanford.edu} (equal advising).}
}

\maketitle

\section{Improvement Strategies Implementation Details}
\label{appendix:strategies}
\subsection{Visual Search}
\label{appendix:visualsearch}
\begin{figure}[t] 
  \centering
  \includegraphics[width=\linewidth]{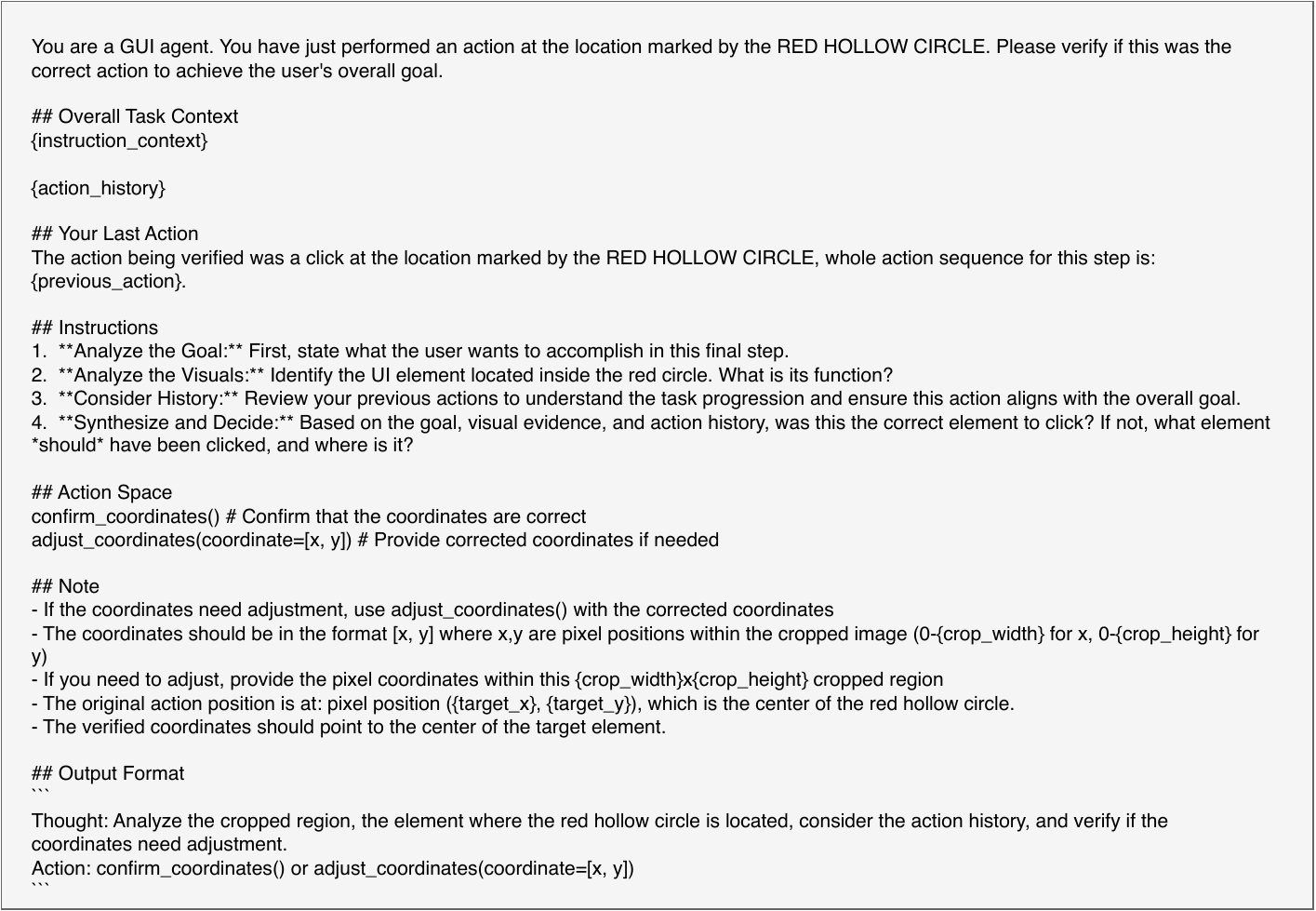} 
  \caption{Prompt for verification process.}
  \label{fig:visual_search_prompt}
\end{figure}
For click-based interactions, we implement a visual search module that highlights the target region before passing it to the verification process. Given a predicted click location, we first crop a \textbf{400$\times$400} image patch centered at the click point. The cropped patch is then \textbf{upscaled by a factor of 2} to provide higher-resolution local context. To explicitly mark the intended target, we draw a \textbf{red hollow circle} of radius \textbf{7 pixels} centered at the click location on the upsampled patch.

To provide the full context of the grounding action and guide the model to reflect on its prior action, we give the agent the task instruction, action history, current click action (with coordinates masked), and the annotated patch, organized as in Fig.~\ref{fig:visual_search_prompt}. The agent can then either confirm the coordinates it produced or adjust the prediction.

If the coordinate is adjusted, we replace the current action with the refined one and proceed to the next step of the task.

\subsection{Terminal Execution}
\label{appendix:terminalexecution}
We provide terminal execution so the agent can perform system-level operations. To encourage tool use, the agent is given the following instruction:

\begin{quote}
\textit{``Tool: Terminal --- Use the keyboard shortcut Ctrl + Alt + T to open a terminal window. Consider this a shortcut for some complex tasks.''}
\end{quote}

This prompt constrains the model to open the terminal only through the designated hotkey. For the \textbf{GUI-Owl-32B} agent, which has limited ability to use terminal commands, we additionally insert a brief instruction on handling password-protected commands and a set of critical terminal rules once the terminal is detected as active. These rules specify that
(1) the terminal must always be opened via the keyboard shortcut \texttt{["ctrl", "alt", "t"]};
(2) before issuing any terminal command, the model must use a \texttt{search} action to verify the correct syntax; and
(3) all command-line operations (\texttt{cd}, \texttt{ls}, \texttt{ln}, \texttt{ffmpeg}, etc.) require an explicit ``Enter'' action to run. This guidance ensures robust and efficient terminal use.

\subsection{Knowledge Support - Search Engine}
\label{appendix:searchengine}
\begin{figure}[t] 
  \centering
  \includegraphics[width=\linewidth]{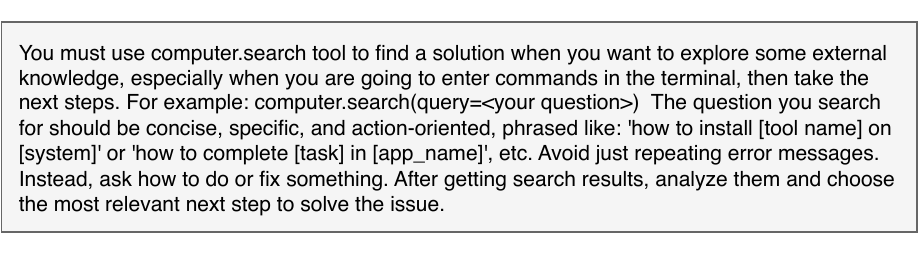} 
  \caption{Prompt for Search Engine.}
  \label{fig:search_engine_prompt}
\end{figure}
For tasks whose solution lies outside the model's inherent capability, we introduce a search-based assistance mechanism inspired by how humans approach computer-use tasks: when users encounter unfamiliar operations, they often consult online resources for the required procedural knowledge. Following this intuition, we allow the agent to issue queries to a proprietary search model, GPT-5-mini, which provides concise, operation-oriented guidance; the prompt template is shown in Fig.~\ref{fig:search_engine_prompt}.

Unlike querying the open web, this controlled search engine returns focused, actionable information and avoids irrelevant or noisy content. To support complex decision-making, the agent may perform \emph{multiple searches within a single step}, iteratively refining its understanding before generating an action. The resulting search outputs are integrated into the action-generation pipeline as an external knowledge supplement that bridges gaps beyond the base model's capability.

\subsection{Knowledge Support - Software Manual}
\label{appendix:softwaremanual}
\begin{table}[tb]
\centering
\footnotesize
\begin{tabularx}{\linewidth}{p{1.5cm} p{2.0cm} p{2.0cm} X}
\toprule
\textbf{App} & \textbf{Hotkey} & \textbf{Function} & \textbf{Detailed Description} \\
\midrule

\multirow{7}{*}{\textbf{Calc}}
& Ctrl + D & Fill Down &
Fill the content or formula of the topmost cell into all other cells within the selected column range. \\
\cline{2-4}

& Ctrl + Enter & Apply to All Selected &
Apply the same formula or value to every cell in the selected range. \\
\cline{2-4}

& Shift + Down & Extend Selection Downward &
Extend selection one cell at a time downward for partial-column fill or formula application. \\
\cline{2-4}

& Ctrl + Shift + Down & Select to Last Nonblank Cell &
Select from the current cell to the last nonblank cell below. \\
\cline{2-4}

& Ctrl + Up / Ctrl + Down & Jump to Top / Bottom of Column &
Move directly to the first or last nonblank cell in a column. \\
\cline{2-4}

& Ctrl + Shift + Right & Select Rightward Range &
Expand the selection horizontally across adjacent columns. \\
\cline{2-4}

& Ctrl + Z & Undo Last Action &
Undo the previous action such as fill, paste, or formula application. \\
\midrule

\multirow{8}{*}{\textbf{Writer}}
& Ctrl + F12 & Insert Table &
Insert a table at the current cursor position. \\
\cline{2-4}

& Ctrl + L & Left Align Paragraph &
Align the current paragraph to the left. \\
\cline{2-4}

& Ctrl + R & Right Align Paragraph &
Align the current paragraph to the right. \\
\cline{2-4}

& Ctrl + E & Center Align Paragraph &
Center-align the current paragraph. \\
\cline{2-4}

& Shift + Right & Select Character &
Select the next character to the right. \\
\cline{2-4}

& Ctrl + Shift + Right & Select Word &
Select the next whole word to the right for quicker editing. \\
\cline{2-4}

& Shift + End & Select to Line End &
Select text from the cursor to the end of the current line. \\
\cline{2-4}

& Ctrl + Shift + Down & Select to Paragraph End &
Select text from the cursor to the end of the current paragraph. \\
\bottomrule
\end{tabularx}
\caption{A consolidated software-manual reference containing curated hotkeys for LibreOffice Calc and LibreOffice Writer.}
\label{tab:hotkey}
\end{table}

In the LibreOffice setting, we additionally provide the agent with a compact software-manual reference tailored to common document-editing operations. Rather than supplying the full official manual, which is lengthy and often irrelevant to the tested tasks, we curate a focused subset of the keyboard shortcuts most frequently needed for GUI-based editing.

This reference sheet (Table~\ref{tab:hotkey}) lists each hotkey together with its functionality (e.g., text formatting, navigation, file operations), offering a lightweight but practical guide that the agent can consult at test time. Grounding the model with this targeted set of actionable hotkeys enables more reliable and deterministic operation of complex software interfaces.

\subsection{Repetition Warnings}
\label{appendix:repetitionwarning}
To prevent the agent from entering unproductive loops, we implement a lightweight repetition-warning module that monitors three types of stagnation signal within a short sliding window.

\noindent
\textbf{Detection rule.}
For thought and action repetition, we use a window of the latest 5 steps and flag a repetition event whenever a pattern appears at least 3 times.
For screen-state repetition, each screen is encoded into a compact representation derived from the OSWorld accessibility tree: we extract the relevant UI-node subtree, compute a stable hash of its structural and textual attributes, and track changes across frames. A warning is triggered only when the hashed accessibility-tree representation remains identical for 3 consecutive steps despite executed actions, indicating that the interface has not progressed.

\noindent
\textbf{Intervention.}
Once a repetition warning is triggered, we append a short instruction to the agent's prompt encouraging it to try an alternative strategy (e.g., issuing a search query, considering a different tool, or switching to terminal operations) rather than persisting in an unproductive loop.

\section{LLM-Guided Self-Improvement Loop Details.}
\subsection{Overview of LLM-Guided Self-Improvement Loop.}
\label{appendix:selfimprovementloop}
To better exploit the information contained in failed trajectories, we incorporate a large language model (LLM) as a test-time diagnostic module.

For each failed rollout produced by the base agent, we feed the LLM a structured input containing the task instruction, action history, and thought process.

Based on this context, the LLM identifies the underlying failure modes and proposes targeted test-time adjustments, which are further classified and implemented as (i) conceptual corrections, (ii) alternative high-level plans, or (iii) direct code-level patches that can be executed before the next rollout. Light human verification is then applied to ensure the robustness of the fixes.

This creates an automatic refinement loop in which improvement is guided by the LLM and failed trajectories are fully leveraged.

\subsection{Full Algorithm.}
\label{appendix:fullalgorithm}
The full refinement procedure is summarized in Algorithm~\ref{alg:self_improvement}. The pipeline comprises three stages:
(i) executing the current agent to obtain a trajectory,
(ii) invoking an LLM to diagnose the failure in context and propose targeted test-time fixes, and
(iii) applying the fix and re-running the task to obtain improved failure experience.
\begin{algorithm}[t]
\caption{Test-Time Self-Improvement via LLM Diagnostics}
\label{alg:self_improvement}
\begin{algorithmic}[1]
\Require Evaluation tasks $\mathcal{T}=\{t_1,\dots,t_N\}$;
initial agent $A_0$; diagnostic LLM $L$; maximum iterations $K$
\Ensure Refined agent $A$

\State $A \gets A_0$
\For{$\text{iter} = 1, \dots, K$}
    \State $\mathcal{D} \gets [\,]$ \Comment{all trajectories}
    \State $\mathcal{F} \gets [\,]$ \Comment{failed trajectories}

    \Statex
    \State \textbf{Run agent and collect trajectories}
    \For{each task $t \in \mathcal{T}$}
        \State $\tau \gets \textsc{Execute}(A, t)$
        \State Append $\tau$ to $\mathcal{D}$
        \If{$\tau$ indicates failure}
            \State Append $\tau$ to $\mathcal{F}$
        \EndIf
    \EndFor

    \If{$\mathcal{F}$ is empty}
        \State \textbf{break} \Comment{early convergence}
    \EndIf

    \Statex
    \State \textbf{LLM-based diagnosis}
    \State $\mathcal{G} \gets [\,]$ \Comment{diagnoses}
    \For{each $\tau \in \mathcal{F}$}
        \State $g \gets L.\textsc{Diagnose}(\tau)$
        \State Append $g$ to $\mathcal{G}$
    \EndFor

    \Statex
    \State \textbf{Propose fixes}
    \State $\mathcal{H} \gets [\,]$ \Comment{fix proposals}
    \For{each diagnosis $g \in \mathcal{G}$}
        \State $h \gets L.\textsc{ProposeFix}(g)$
        \State Append $h$ to $\mathcal{H}$
    \EndFor

    \Statex
    \State \textbf{Optional human verification}
    \State $\mathcal{H} \gets \textsc{FilterValid}(\mathcal{H})$

    \Statex
    \State \textbf{Refine the agent}
    \State $A \gets \textsc{ApplyFixes}(A, \mathcal{H})$
\EndFor

\State \Return $A$
\end{algorithmic}
\end{algorithm}

\subsection{Example: Correcting a Font-Change Failure with Action Loop Detection.}
\label{appendix:example}

This section illustrates how Algorithm~\ref{alg:self_improvement} operates on a real failure case from LibreOffice Impress. The task is

\begin{quote}
\textit{``Change all fonts in the presentation to Liberation Sans Narrow.''}
\end{quote}

\paragraph*{Initial Failure}
The base agent repeatedly enters \textit{Master View}, applies a font change,
exits, and re-enters \textit{Master View}. This cycle continues for over
20 iterations until the 30-step limit is reached. The LLM then provides a detailed analysis of the failure mechanism:
\begin{itemize}
    \item \textbf{Failure type}: infinite loop with view transitions.
    \item \textbf{Root cause}: font edits in \textit{Master View} do not
    propagate to normal slides.
    \item \textbf{Missing mechanism}: lack of loop detection and success
    verification.
\end{itemize}

\paragraph*{LLM-Proposed Fix: Loop Detection Module}
The LLM proposes adding a loop detector that is triggered when the same high-level
action pattern appears repeatedly:

\begin{lstlisting}[language=Python]
def _detect_action_repetition(
    self, current_action_text: str,
    lookback_window: int = 5,
    repetition_threshold: int = 2)
    -> tuple[bool, int]:
    
    if not self.enable_action_
        repetition_detection:
        return False, 0

    if not self.actions 
        or len(self.actions) == 0:
        return False, 0

    # Get the last N actions 
    # (up to lookback_window)
    recent_actions = (
        self.actions[-lookback_window:] 
        if len(self.actions) 
            >= lookback_window 
        else self.actions
    )
\end{lstlisting}

A light human check confirms that the patch is safe, and the test-time hook is implemented.

The action-loop-detection module is thus integrated into the framework, and another round of the rollout-and-diagnosis loop is executed.

\subsection{Human Involvement in the Failure-Driven Loop}
\label{sec:human}
In this section, we clarify the extent of human involvement in our framework. In the failure-driven loop, an LLM judge first analyzes failed trajectories, identifies common failure modes, and proposes candidate solutions. A human then selects among these options, after which the LLM automatically generates the corresponding code patch.

In practice, the generated patches are largely correct, and human intervention is typically limited to minor syntactic or runtime fixes. Quantitatively, such adjustments account for less than 3\% of the modified lines on average, indicating that the overall process remains largely automated. Human involvement is therefore minimal rather than substantial engineering.

\end{document}